\documentclass{article}

   \usepackage[preprint]{neurips_2023}

\usepackage[utf8]{inputenc} 
\usepackage[T1]{fontenc}    
\usepackage{hyperref}
\usepackage{url}            
\usepackage{booktabs}       
\usepackage{amsfonts}       
\usepackage{nicefrac}       
\usepackage{microtype}      
\usepackage{xcolor}         

\usepackage{amsmath}
\usepackage{amssymb}
\usepackage{mathtools}
\usepackage{amsthm}

\usepackage{blindtext}
\usepackage{multicol}
\usepackage{color}
\usepackage[capitalize,noabbrev]{cleveref}

\theoremstyle{plain}

\theoremstyle{definition}

\theoremstyle{remark}

\usepackage{amsmath,amsfonts,bm}

\def\eqref#1{equation~\ref{#1}}

\def\1{\bm{1}}

\DeclareMathAlphabet{\mathsfit}{\encodingdefault}{\sfdefault}{m}{sl}
\SetMathAlphabet{\mathsfit}{bold}{\encodingdefault}{\sfdefault}{bx}{n}

\DeclareMathOperator*{\argmin}{arg\,min}

\newcommand{\independent}{\perp \!\!\! \perp}

\usepackage{refcount}
\usepackage{url}
\usepackage{float}
\usepackage{xcolor,colortbl}
\definecolor{Gray}{gray}{0.85}
\definecolor{LightCyan}{rgb}{0.85,0.92,0.90}

\usepackage{subfig}
\usepackage{xspace}
\usepackage{graphicx}
\usepackage{caption}

\usepackage{multirow}
\usepackage{enumitem}

\usepackage{pifont}
\newcommand{\cmark}{\ding{51}}%
\newcommand{\xmark}{\ding{55}}%

\usepackage{booktabs}

\title{Probabilistic Self-supervised Learning via \\
Scoring Rules Minimization}

\author{Amirhossein Vahidi \\
LMU Munich \& \\  MCML 
  \And
Simon Schoßer  \\
 LMU Munich
\And
Lisa Wimmer \\
LMU Munich \& \\  MCML 
\And
Yawei Li \\
LMU Munich \& \\  MCML 
  \And
Bernd Bischl \\
LMU Munich \& \\  MCML 
\And
Eyke Hüllermeier \\
LMU Munich \& \\ MCML 
\And
Mina Rezaei \\
LMU Munich \& \\ MCML 
}

\begin{document}

\maketitle

\begin{abstract}
In this paper, we propose a novel probabilistic self-supervised learning via Scoring Rule Minimization (ProSMIN), which leverages the power of probabilistic models to enhance representation quality and mitigate collapsing representations.
Our proposed approach involves two neural networks; the online network and the target network, which collaborate and learn the diverse distribution of representations from each other through knowledge distillation. By presenting the input samples in two augmented formats, the online network is trained to predict the target network representation of the same sample under a different augmented view. The two networks are trained via our new loss function based on proper scoring rules. We provide a theoretical justification for ProSMIN's convergence, demonstrating the strict propriety of its modified scoring rule. This insight validates the method's optimization process and contributes to its robustness and effectiveness in improving representation quality.
We evaluate our probabilistic model on various downstream tasks, such as in-distribution generalization, out-of-distribution detection, dataset corruption, low-shot learning, and transfer learning. Our method achieves superior accuracy and calibration, surpassing the self-supervised baseline in a wide range of experiments on large-scale datasets like ImageNet-O and ImageNet-C, ProSMIN demonstrates its scalability and real-world applicability.
\end{abstract}

\section{Introduction}

In the field of machine learning, probabilistic approaches have emerged as a powerful toolkit, offering advantages such as uncertainty quantification, robustness, and flexibility, especially in domains with inherent complexity and uncertainty~\cite{dutordoir2021deep,peharz2020einsum,ghahramani2015probabilistic}. While deterministic models have their merits, a crucial frontier lies in extending the benefits of probabilistic methods to self-supervised learning (SSL) for representation reliability. Representation reliability provides a more comprehensive understanding of the underlying data, which is particularly valuable in domains where prediction errors can have serious consequences, such as disease diagnosis~\cite{ozdemir20193d,rezaei2022bayesian}, climate prediction~\cite{gneiting2007strictly}, computational finance~\cite{dawid2014theory}, and economic forecasting~\cite{gneiting2014probabilistic}.

Current SSL methods have made impressive progress, but a notable gap exists in their evaluation of representational reliability through a probabilistic lens. Incorporating probabilistic modeling into SSL training provides a way to not only predict outcomes but also provide distributions of possible predictions along with associated probabilities~\cite{ho2020denoising,nichol2021improved}. This distinctive feature enables SSL models to capture the inherent uncertainties in real-world data and provide a more complete understanding of underlying patterns. Encouragingly, recent advances in SSL, exemplified by methods such as \cite{oquab2023dinov2}, have demonstrated the potential to generate effective features across diverse image distributions and tasks without the need for fine-tuning, confirming the viability of this paradigm shift.

As the demand for more reliable representations escalates, the integration of probabilistic principles into SSL is well poised to improve representation quality, align predictions with real-world complexity, and thereby advance the frontiers of machine learning. In this paper, we present ProSMIN, a novel probabilistic formulation of self-supervised learning that contributes to the understanding and improvement of probabilistic modeling in this domain. Our proposed method involves two deep neural networks, an online and a target network, each learning a different representation of input samples. The online network maps input samples to a probability distribution inferred from the representation of the online encoder part. We train the online network in such a manner that samples from its output distribution predict the target network's representation on a second augmented view of the input. The loss realized by this prediction is expressed via a modified scoring rule, which incentivizes the recovery of the true distribution.

Our contributions are:
\begin{itemize}
    \item We introduce a novel probabilistic definition of representation reliability for self-supervised learning. The probabilistic definition provides a deeper understanding of the quality and trustworthiness of the learned representations in guiding subsequent tasks. To the best of our knowledge, this is the first comprehensive study to investigate probabilistic formulation in the self-supervised representation domain.
    \item Our probabilistic approach effectively mitigates collapsing representation by encouraging the online and target networks to explore a diverse range of representations, thus avoiding convergence to a limited set which results in more comprehensive representation space that better encapsulates the intricacies of the data distribution.
    \item We provide a rigorous theoretical foundation for the convergence of our proposed algorithm. Specifically, we demonstrate the strict propriety of our modified scoring rule, substantiating the convergence of the optimization process. This theoretical insight not only underscores the robustness of our approach but also provides a principled explanation for its effectiveness in improving representation quality.
    \item Through extensive empirical analysis, we validate the effectiveness of our approach in diverse scenarios. Our method achieves competitive predictive performance and calibration on various tasks such as in-distribution (IND), out-of-distribution (OOD), and corrupted datasets, demonstrating generalization capabilities. Moreover, we demonstrate the superiority of our method in semi-supervised and low-shot learning scenarios. Additionally, our framework establishes a superior trade-off between predictive performance and robustness when compared to deterministic baselines. This outcome is particularly notable on large-scale datasets such as ImageNet-O and ImageNet-C, underscoring the scalability and effectiveness of our method in real-world, high-dimensional settings.
\end{itemize}

\section{Background and Related work}

Self-supervised methods are designed to tackle unsupervised problems by training on a \emph{pretext task} that utilizes the data itself to generate labels, effectively employing supervised methods to solve unsupervised problems~\cite{grill2020bootstrap,chen2020simple,jang2023self,caron2021emerging,zhou2021ibot,zbontar2021barlow,bardes2021vicreg,chen2021empirical}. The resulting representations learned from the pretext task can serve as a foundation for \emph{downstream supervised tasks}, such as image classification or object detection. Alternatively, the extracted representation can be directly utilized for downstream applications, such as detecting anomalies and OOD data~\cite{tran2022plex}. Recent studies~\cite{oquab2023dinov2,zhou2021ibot} provided evidence that performing self-supervised pretext task learning on a large-scale and diverse dataset can extract features that are effective across different image distributions and tasks without the need for fine-tuning. Following this, we introduce a novel probabilistic self-supervised framework aiming to learn robust representation over parameters using \emph{self-distillation} and by \emph{minimizing the scoring rule}.

Self-distillation is a variant of knowledge distillation~\cite{hinton2015distilling} in which a larger model (\emph{teacher}) is used to distill knowledge into a smaller model (\emph{student}) of the same architecture~\cite{caron2021emerging,zhou2021ibot}. Given an input sample $\bm{x}$, the student network $f_{\bm{\theta}}$ is trained on the soft labels provided by the teacher network $f_{\bm{\xi}}$. Self-distillation combines self-supervised learning with knowledge distillation and was introduced by DINO~\cite{caron2021emerging}. The two networks share the same architecture but take different augmentations of the input sample and output different representation vectors. The knowledge is distilled from the teacher network $f_{\bm{\xi}}$ to student $f_{\bm{\theta}}$ by minimizing the cross-entropy between the respective representation vectors. The parameters of the student network $\bm{\theta}$ are obtained from an exponential moving average (EMA) of the parameters of the teacher network $\bm{\xi}$, thus reducing computational cost by confining backpropagation to the teacher network.

A scoring rule is a function used to evaluate the accuracy of a probabilistic prediction~\cite{gneiting2007strictly,Kiureghian2009}. It quantifies the divergence between the predicted probability distribution and the true distribution of the event. The concept of a scoring rule is fundamental to many areas of machine learning, including probabilistic classification~\cite{parry2016linear} and decision theory~\cite{dawid2014theory}. A proper scoring rule~\cite{gneiting2014probabilistic} is one that incentivizes truthful reporting of the probabilities by the forecaster, i.e., the forecaster is incentivized to report the correct probability distribution~\cite{pacchiardi2021probabilistic}. The use of proper scoring rules has had significant implications in areas such as online learning~\cite{v2019online}, generative neural networks~\cite{pacchiardi2022likelihood}, and uncertainty quantification~\cite{bengs2023second,gruber2022better}. This paper utilizes scoring rules and adapts them for the purpose of pretext task learning of a self-supervised framework. This adaptation is inspired by the endeavor to infuse probabilistic learning principles into the self-supervised learning domain.

\section{Problem formulation}

\paragraph{Avoiding collapsing representation}
One of the major challenges in self-supervised learning is collapsing representation where learned representations converge to a limited set of points in the representation space. In other words, the model fails to capture the full diversity and richness of the underlying data distribution. This can lead to reduced representation quality, impaired generalization to new tasks or domains, and limited capacity to handle variations in the data. Most recent studies addressed this problem with contrastive learning through effective augmentation~\cite{chen2020simple}, negative sample strategies~\cite{wang2021instance}, ensemble approach~\cite{vahidi2023diversified}, regularization techniques~\cite{rezaei2023deep}, and self-distilation~\cite{caron2021emerging}. In this paper, we formulate the collapsing representation problem through a probabilistic lens, aiming to provide a comprehensive and nuanced solution that not only addresses the limitations of deterministic approaches but also harnesses the power of uncertainty quantification and broader representation distributions. This pivotal shift offers promising avenues for achieving representation reliability and superior generalization capabilities in self-supervised learning scenarios.

\paragraph{Scoring rules}
A scoring rule~\cite{gneiting2007strictly} is a function that evaluates how well a predicted distribution $P$ over a random variable $\mathbf{X}$ aligns with the actually observed realizations $\bm{x}$ of $\mathbf{X}$.
We define the loss\footnote{
The original proposal in \cite{gneiting2007strictly} defines scoring rules in terms of a gain that is to be maximized. We adhere to the convention in deep learning of expressing the objective via a loss function that we seek to minimize.
} of predicting distribution $P$ while observing $\bm{x}$ as $S(P, \bm{x})$.
Assuming that $\mathbf{X}$ follows some true distribution $Q$, the \emph{expected} scoring rule measuring the loss of predicting $P$ can be expressed as $S(P, Q) \triangleq \mathbb{E}_{\mathbf{X} \sim Q} S(P, \mathbf{X})$.
A scoring rule $S$ is \emph{proper} with respect to a set of distributions $\mathcal{P}$ if for all $P, Q \in \mathcal{P}$ it holds that $S(Q, Q) \leq S(P, Q)$, thus incentivizing the prediction of the true distribution $Q$. 
If the former holds with equality, i.e., the expected score $S(P, Q)$ is uniquely minimized in $Q$ at $Q = P$, then the scoring rule is called \emph{strictly proper}. 
In practice, the expectation with respect to $Q$ is usually replaced by an empirical mean over a finite amount of samples.
We will denote the resulting scoring rule as $\hat{S}$. 
There exist many types of scoring rules, including entire parameterized families of strictly proper scoring rules. We provide more details on some of them that we use in our experiments in Appendix~\ref{sec:scoring_rule}.

\begin{figure}
\centering
    \includegraphics[width=0.9\textwidth]{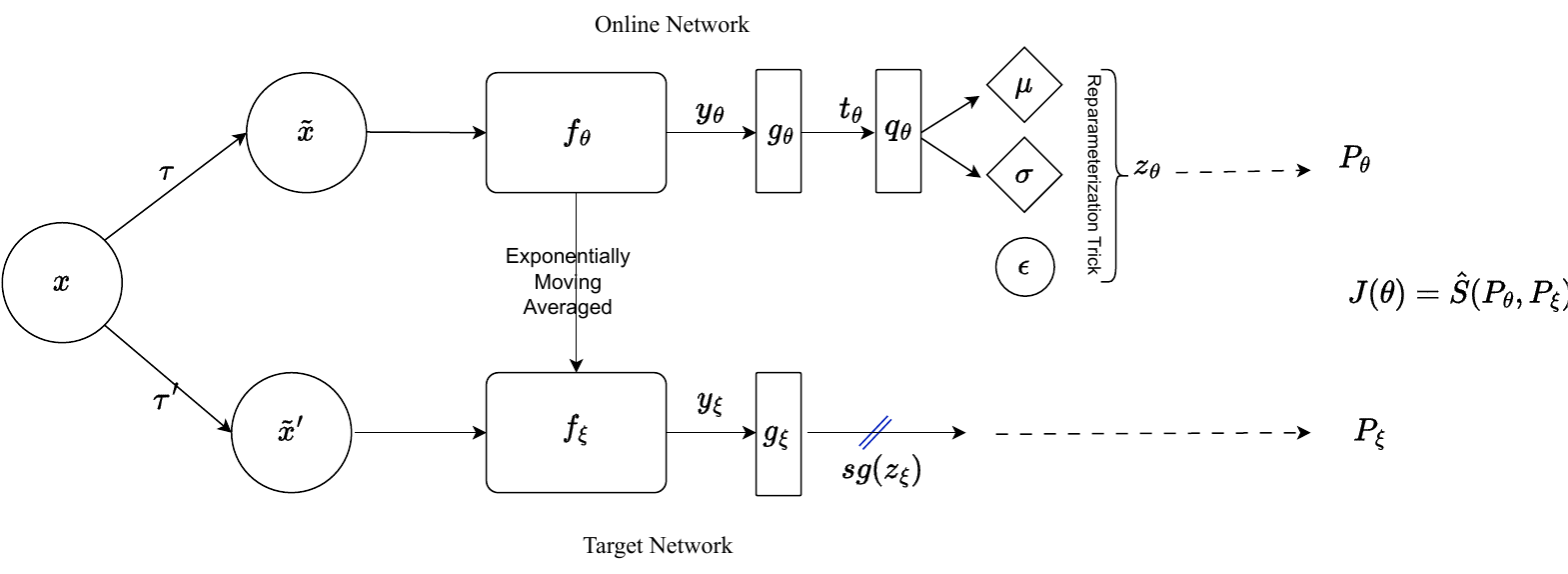}
    \caption{Illustration of our proposed architecture (ProSMin). Given a batch $\bm{X}$ of input samples, two different augmented samples $\tilde{\bm{x}}$ and $\tilde{\bm{x}}^\prime$ are taken by an online network with $\bm{\theta}$ and target network with $\bm{\xi}$ parameters respectively. Our objective is to minimize the proposed scoring rule between $P_{\bm{\theta}}$, $P_{\bm{\xi}}$.}\label{fig:method}
\end{figure}



\section{Method}

Consider a randomly sampled mini-batch of training data $\boldsymbol{X} \triangleq \left[\boldsymbol{x}_1, \ldots, \boldsymbol{x}_n\right]^N \in \mathbb{R}^{N \times D}$ and transformation functions $\tau, \tau^\prime$ acting on the data. To enhance the training process, the transformation functions produce two augmented views $\tilde{\bm{x}} \triangleq \tau(\bm{x})$ and $\tilde{\bm{x}}^\prime \triangleq \tau^\prime(\bm{x})$ for each sample in $\bm{X}$. These augmented views are generated by sampling $\tau, \tau^\prime$ from a distribution of suitable data transformations, such as partially masking image patches~\citep{he2022masked} or applying image augmentation techniques~\citep{chen2020simple}.

The first augmented view $\tilde{\bm{x}}$ is fed to the encoder of online network $f_{\bm{\theta}}$ that outputs a \emph{representation} $y_{\bm{\theta}} \triangleq f_{\bm{\theta}} (\tilde{\bm{x}})$. This is followed by passing it subsequently through a projector $g_{\bm{\theta}}$ and predictor $q_{\bm{\theta} }$, such that $t_{\bm{\theta}} \triangleq g_{\bm{\theta}}(y_{\bm{\theta}})$ and $z_{\bm{\theta}} \triangleq q_{\bm{\theta} }(t_{\bm{\theta}})$.
We collect all trainable parameters of the online network in $\bm{\theta}$. Similarly, the encoder of target network $f_{\bm{\xi}}$ takes the second augmented view and outputs $y_{\bm{\xi}} \triangleq f_{\bm{\xi}}(\tilde{\bm{x}}^\prime)$, followed by the projector network $g_{\bm{\xi}}$ producing $t_{\bm{\xi}} \triangleq g_{\bm{\xi}}(y_{\bm{\xi}})$, where the trainable parameters are denoted as $\bm{\xi}$. It is important to note that the predictor is applied exclusively to the online network.

In order to introduce probabilistic self-distillation training, we employ a scoring rule~\cite{gneiting2007strictly} as our loss function. To accomplish this, it is necessary to generate samples from the online network. 
One way of producing samples from a neural network architecture, while still enabling backpropagation with respect to the latent representation $\bm{z}$, is to use the reparametrization trick~\cite{kingma2015variational}. Here, we assume that the output of the predictor network $z_{\bm{\theta}} = q_{\bm{\theta} }(t_{\bm{\theta}})$ follows an underlying normal distribution with mean $\mu$ and standard deviation $\sigma$. We generate $r \in \mathbb{N}$ samples from the output of the linear layers following the prediction head by sampling random noise $\epsilon^{i}_{j} \sim N(0,1)$ for each augmented view of the $i$-th data point, such that the $j$-th sample, with $j \in 1,...,r$, is given by: $\bm{z}^{i}_{j} = \mu + \sigma \odot  \epsilon^{i}_{j}.$  


We thus obtain samples $\bm{z}^{i}_{j}$ by shifting and scaling the random noise samples $\epsilon^{i}_{j}$ by the outputs $(\mu, \sigma)$ of a neural network with trainable parameters $\bm{\theta}$, and the loss incurred by these samples can be backpropagated to update $\bm{\theta}$ during training. 
We update the online network parameters by minimizing the scoring rule as follows:

\begin{equation}\label{eq:1}
    \hat{\bm{\theta}}:=\argmin_{\bm{\theta}} J(\bm{\theta}), \quad J(\bm{\theta}) =\hat{S}(P_{\bm{\theta}}, P_{\bm{\xi}}):=\mathbb{E}_{z_{\bm{\xi}} \sim P_{\bm{\xi}}} [P_{\bm{\theta}}, z_{\bm{\xi}}],
\end{equation}

where $P_{\bm{\xi}}$ denotes the target distribution and $z_{\bm{\xi}}$ denotes the target output and $P_{\bm{\theta}}$ represents the online induced multivariate normal distribution. 
Our custom scoring rule loss over $P_{\bm{\theta}}$ and $P_{\bm{\xi}}$ is defined as

\begin{equation}~\label{eq:2}
    \hat{S}(P_{\bm{\theta}}, P_{\bm{\xi}}) = 
    \dfrac{1}{N} \sum_{i=1}^{N} \left| \!\left[ 
    \dfrac{2 \lambda}{r} \sum_{j=1}^{r} \Vert \bm{z}_{j}^{i} - \bm{z}_\xi^{i}\Vert^{\beta}_{2} -
    \frac{1- \lambda}{r(r-1)} \sum_{j \neq k} \Vert \bm{z}_{j}^{i} - \bm{z}_{k}^{i} \Vert^{\beta}_{2} \!\right] \right|,
\end{equation}

where $\bm{z}_{\xi}^{i}$ represents the target prediction for the $i$-the input sample. $\beta \in (0,2)$ and $\lambda \in (0,1)$ are hyperparameters. Detailed information is provided in Appendix~\ref{Implementation}.

By the principle of knowledge distillation, the parameters of the target network are updated through the EMA of the weights from the online network~\cite{grill2020bootstrap,caron2021emerging}, saving the need for backpropagation and thus reducing computation time considerably.

\begin{align}
    \bm{\xi}_t &= \alpha \bm{\theta}_t + (1-\alpha)\bm{\xi}_{t-1}, \quad t = 1, 2, \dots, \quad \alpha \in [0,1]
\end{align}

The initial weights $\bm{\xi}_0$ are obtained through random initialization.

Next, we explain the details of our objective function based on scoring rules~\ref{sec:method:loss}. We refer to Section~\ref{sec:scoring_rule} for the background and different variations of scoring rules. Then, we show that, by utilizing a strictly proper scoring rule and retaining the ability to calculate gradients as usual (as proven in \ref{bem_exchange}), we can infer that our algorithm will converge towards the desired minimum (see Section~\ref{sec:method:proof}).

\subsection{Avoiding collapse}~\label{sec:method:collapse}
There are different methods in self-supervised learning to avoid collapse by adding batch normalization \cite{grill2020bootstrap}, prediction layer \cite{grill2020bootstrap}, and sharpening and centering \cite{caron2021emerging}. Our framework can benefit from more layers in the online network and centering on the target outputs to improve the results. Still, our method inherently can not collapse to the same representation because of its probabilistic nature. Furthermore, our loss function converges to zero without collapsing.

\subsection{Objective function}~\label{sec:method:loss}
Many scoring rules decompose into two terms, one of which is a function of two samples $\bm{z}_j, \bm{z}_k$ from the predicted distribution $P$, while the other is a function of $\bm{z}_j$ and the realized observation $\bm{z}_{\bm{\xi}}$.
Our objective function is an adjusted version of the former, where the two components are posed as a convex combination with component weights controlled via hyperparameter $\lambda \in (0,0.5]$.
This adjustment of the scoring rules can be useful to adjust the focus of the loss function on either part of the scoring rule.
The hyperparameter $\lambda$ helps to improve the performance by shifting the focus of the loss function.
Two notable examples of scoring rules adhering to this form are the \emph{energy score} and the \emph{kernel score}~\cite{gneiting2007strictly}.

With the above notation, we define the energy score as:

\begin{equation}~\label{eq:energyloss}
    S_{\mathrm{E}}(P_{\bm{\theta}}, \bm{z}_{\bm{\xi}})=2 \cdot \mathbb{E}_{P_{\bm{\theta}}}\left[\|\bm{z}_j-\bm{z}_{\bm{\xi}}\|_2^\beta\right]-\mathbb{E}_{P_{\bm{\theta}}}\left[\left\|\bm{z}_j-\bm{z}_k \right\|_2^\beta\right]
    =: S_{\mathrm{E}}^{1}(P_{\bm{\theta}}, \bm{z}_{\bm{\xi}}) + S_{\mathrm{E}}^{2}(P_{\bm{\theta}}, \bm{z}_{\bm{\xi}})
\end{equation}

Analogously, we write the kernel score as:


\begin{equation}~\label{eq:energy}
    S_{\mathrm{K}}(P_{\bm{\theta}}, \bm{z}_{\bm{\xi}})= \mathbb{E}_{P_{\bm{\theta}}}\left[k(\bm{z}_j, \bm{z}_k )\right] -
    2 \cdot \mathbb{E}_{P_{\bm{\theta}}}\left[k(\bm{z}_j, \bm{z}_{\bm{\xi}})\right]
    =: S_{\mathrm{K}}^{2}(P_{\bm{\theta}}, \bm{z}_{\bm{\xi}}) + S_{\mathrm{K}}^{1}(P_{\bm{\theta}}, \bm{z}_{\bm{\xi}}),
\end{equation}

with suitable kernel function $k(\cdot, \cdot)$.

For simplicity, we only write $S(P_{\bm{\theta}}, \bm{z}_{\bm{\xi}}) = S^{1}(P_{\bm{\theta}}, \bm{z}_{\bm{\xi}})+S^{2}(P_{\bm{\theta}}, \bm{z}_{\bm{\xi}})$ in the following for both scores. With $\lambda \in (0,1)$, we define the general form of our objective function as follows:

\begin{equation}
    S^{*}(P_{\bm{\theta}}, \bm{z}_{\bm{\xi}}) := \left| \lambda S^{1}(P_{\bm{\theta}}, \bm{z}_{\bm{\xi}}) + (1-\lambda) S^{2}(P_{\bm{\theta}}, \bm{z}_{\bm{\xi}}) \right|.
\end{equation}

The proposed objective function for energy and kernel score is strictly proper in our setup.
We provide proof in following Section~\ref{sec:theorey}. 

\subsection{Theoretical justification} ~\label{sec:theorey}
This section provides the mathematical justification of our algorithm. 
We establish that the custom loss is strictly proper in our setup. 
In subsection \ref{bem_exchange}, we demonstrate that the expectation and gradient of our theoretical derivative can be interchanged, thereby enabling us to derive the gradients as usual. 
Due to the use of the samples within the scoring rule as well as the use of non-differential activation functions, this step becomes necessary.
This is followed by subsection \ref{unb_est}, where we present an unbiased estimate of the gradient. 
By undertaking these three steps, we ensure that our algorithm converges effectively.

\paragraph{Strict propriety of the proposed objective function} ~\label{sec:method:proof}
Let $\lambda \in (0,1)$.
We only show the proof for the energy score. The proof for the kernel score can be done analogously.
We need to show that 
\begin{equation}
    S^{*}(P_{\bm{\xi}}, P_{\bm{\xi}}) < S^{*}(P_{\bm{\theta}}, P_{\bm{\xi}}) \quad \forall P_{\bm{\theta}},P_{\bm{\xi}} \in \mathcal{P} \quad s.t. \quad P_{\bm{\theta}} \neq P_{\bm{\xi}}
\end{equation}

We define the following based on Eq. \ref{eq:1}: 
\begin{align*}
    S^{*}(P_{\bm{\theta}}, P_{\bm{\xi}})&= \left| \mathbb{E}_{\bm{z}^i_{\bm{\xi}} \sim P_{\bm{\xi}}} S^{*}(P_{\bm{\theta}}, \bm{z}^i_{\bm{\xi}})\right| \\
    &=\left|\mathbb{E}_{\bm{z}^i_{\bm{\xi}} \sim P_{\bm{\xi}}} \left[\lambda \mathbb{E}_{\bm{z}^i_j \sim P_{\bm{\theta}}} \left[ 2 \|\bm{z}^i_j-\bm{z}^i_{\bm{\xi}}\|_2^\beta\right]-(1-\lambda)\mathbb{E}_{\bm{z}^i_j, \bm{z}^i_k \sim P_{\bm{\xi}}}\left[\left\|\bm{z}^i_j-\bm{z}^i_k\right\|_2^\beta\right] \right]\right| \\
    &> \left|\mathbb{E}_{\bm{z}^i_{\bm{\xi}} \sim P_{\bm{\xi}}} \left[\lambda \mathbb{E}_{\bm{z}^i_j \sim P_{\bm{\xi}}} \left[ 2 \|\bm{z}^i_j-\bm{z}^i_{\bm{\xi}}\|_2^\beta\right]-(1-\lambda)\mathbb{E}_{\bm{z}^i_j, \bm{z}^i_k \sim P_{\bm{\xi}}}\left[\left\|\bm{z}^i_j-\bm{z}^i_k\right\|_2^\beta\right] \right] \right| \\
    &= S^{*}(P_{\bm{\xi}}, P_{\bm{\xi}}) = 0
\end{align*}
$S^{*}(P_{\bm{\xi}}, P_{\bm{\xi}})$ can be estimated as zero and hence is smaller than $S^{*}(P_{\bm{\theta}}, P_{\bm{\xi}})$ for every $P_{\bm{\theta}} \neq P_{\bm{\xi}}$.
For a specific observation $\bm{x}_i$, there is only a single target prediction $\bm{z}_{\bm{\xi}}$, such that $\bm{z}_{\bm{\xi}}$ essentially follows a Dirac distribution.
This concludes the proof.

By utilizing a strictly proper scoring rule and retaining the ability to calculate gradients as usual (as proven in \ref{bem_exchange}), we can infer that our algorithm will converge towards the desired minimum.



\section{Implementation details and experimental setup}~\label{sec:implementation}
\noindent\textbf{Image augmentation}
We define a random transformation function $\bm T$ that applies a combination of multi-crop, horizontal flip, color jittering, and grayscale. Similar to~\cite{caron2021emerging}, we perform multi-crops with a random size from $0.8$ to $1.0$ of the original area and a random aspect ratio from $3/4$ to $4/3$ of the original aspect ratio. We define color-jittering of $(0.8, 0.8, 0.8, 0.2)$, and Gaussian blurring with $0.5$ probability and $\zeta = (0.1, 2.0)$. 

\noindent\textbf{Deep self-supervised network architecture ~}
The \emph{online neural network} is constructed from a backbone $f$, which can be either ViT~\cite{dosovitskiy_2021_ImageWorth16x16} or ResNet~\cite{he2016deep}, and a projection head $g$ followed by a prediction $q$. The backbone $f$ output is used as features for downstream tasks. The projection consists of a 3-layer multilayer perceptron (MLP) with a hidden dimension of 2048, followed by 2 normalizations and a weight-normalized fully connected layer with $K$ dimensions, similar to the design used in the DINO projection head. We use a predictor with two layers of MLPs with a hidden dimension of 12000, with nonlinearity GELU in between. Notably, ViT architectures do not use batch normalization (BN) by default, so we do not use BN in the projection or prediction when applying ViT.

The \emph{target network} has the same backbone and projection as the online network and it learns through self-distillation. Similar to DINO~\cite{caron2021emerging}, after the online network parameters are updated, an EMA of the online parameters (i.e., a momentum encoder) is used to update the target parameters. The EMA prevents the target parameters from updating too quickly. After parameter updates the target also gets a new centering parameter.

\noindent\textbf{Optimization ~}
Our pretraining process involves training the models on the ImageNet training dataset~\cite{deng2009imagenet} using the \textit{adamw} optimizer~\cite{loshchilov2017decoupled} and a batch size of 512, distributed across 8 GPUs utilizing Nvidia Tesla A100 with ViT-S/16 architecture. We adopt a linear scaling rule to determine the base value of the learning rate, which is ramped up linearly during the first 30 epochs. Specifically, the learning rate is set to $lr = 0.0005 * \text{batchsize}/256$. After the warmup phase, we decay the learning rate using a cosine schedule~\cite{loshchilovH1}. The weight decay also follows a cosine schedule, decreasing from $0.04$ to $0.4$. The centering (smooth parameter) is $0.9$.

\noindent\textbf{Datasets ~}
The datasets utilized in our experiments are as follows:
The \textbf{ImageNet}~\cite{deng2009imagenet} dataset with 1.28 million training images and 50,000 validation images with the size of $256 \times 256$ contains 1,000 classes. The \textbf{ImageNet-O} dataset~\cite{srivastava2022out} comprises images belonging to classes that are not present in the ImageNet-1k dataset. It is considered a challenging benchmark for evaluating the robustness of the models, as it requires models to generalize to a diverse range of visual conditions and handle variations that are not typically encountered in standard training datasets.
The \textbf{ImageNet-C}~\cite{HendrycksD19imagenetc} is a benchmark for evaluating the robustness of the models against common corruptions and perturbations that can occur in real-world scenarios. It consists of more than 30,000 images derived from the ImageNet dataset, with each image being corrupted in one of 15 different ways, including noise, blur, weather conditions, and digital artifacts. \textbf{CIFAR-10/100}~\cite{Krizhevsky2009learning} are subsets of the tiny images dataset. Both datasets include 50,000 images for training and 10,000 validation images of size $32\times32$ with 10 and 100 classes, respectively. The \textbf{Oxford 102 Flower}~\cite{nilsback2008automated} consists of 102 flower categories,  each class including between 40 and 258 images. The images have large scale, pose, and light variations. In addition, there are categories that have large variations within the category and several very similar categories. \textbf{iNaturalist-2018}~\cite{van2018inaturalist} (iNat) comprises a vast collection of 675,170 training and validation images, classified into 5,089 distinct fine-grained categories found in the natural world. It is worth noting that the iNat dataset exhibits a significant imbalance, as the number of images varies greatly across different categories.

\noindent\textbf{Tasks ~} 
We evaluate the performance of ProSMin representations after self-supervised pretraining on the ImageNet on the basis of \textbf{In-Domain (IND) generalization}, \textbf{OOD detection}, \textbf{semi-supervised learning}, \textbf{corrupted dataset evaluation}, as well as \textbf{transfer learning to other datasets and tasks}.

\noindent\textbf{Evaluation metrics ~} 
We report the prediction performance with the following metrics: \textbf{Top-1 accuracy $\uparrow$}: refers to the proportion of test observations that are correctly predicted by the model's output as belonging to the correct class.
\textbf{AUROC} $\uparrow$: the area under the ROC curve represents the relationship between false-positive and false-negative rates for various classification thresholds. In this case, the positive and negative classes refer to whether an observation is in or out of a given distribution, respectively, and the ROC curve is plotted as the threshold for classifying an observation as "positive" is gradually increased.
\textbf{Negative log-likelihood (NLL)} $\downarrow$: measures the probability of observing the given test data given the estimated model parameters, multiplied by -1. This measure quantifies the degree to which the model's estimated parameters fit the test observations.
\textbf{Expected calibration error (ECE)} $\downarrow$ \citep{naeini_2015_ece}: calculated as the mean absolute difference between the accuracy and confidence of the model's predictions, where confidence is defined as the highest posterior probability among the predicted classes. The difference is calculated across equally-spaced confidence intervals or bins and is weighted by the relative number of samples in each bin. A lower value of ECE indicates better calibration of the model.
\textbf{mean Calibration Error (mCE)} $\downarrow$ is a metric used to evaluate the calibration of a classification model, similar to ECE. It is calculated as the mean of the absolute differences between the predicted and true probabilities of a given class across all classes. A lower value of mCE indicates better calibration of the model.



\vspace{-10pt}
\section{Results and discussion}\label{sec:results} 
\vspace{-10pt}


\textbf{In-distribution generalization ~}~\label{sec:results:ind}
IND generalization (or \textit{linear evaluation}) measures how well a model's confidence aligns with its accuracy. To assess and compare the predictive abilities of our proposed model on in-distribution datasets, we freeze the encoder of the online network, denoted as $f_{\bm{\theta}}$, after performing unsupervised pretraining. Then, we train a supervised linear classifier using a fully connected layer followed by softmax, which is placed on top of $f_{\bm{\theta}}$ after removing the projection and prediction network. The desired outcome is high predictive scores and low uncertainty scores. 
In Table~\ref{table:calibration:imagenet}, a comprehensive comparison is presented between our approach and other self-supervised methods. The results indicate that our method attained the second-highest accuracy with 300 epochs while outperforming all others in terms of calibration, as evidenced by the lowest Expected Calibration Error (ECE) and Negative Log-Likelihood (NLL) scores.

\begin{figure}
\centering
    \includegraphics[width=0.78\textwidth]{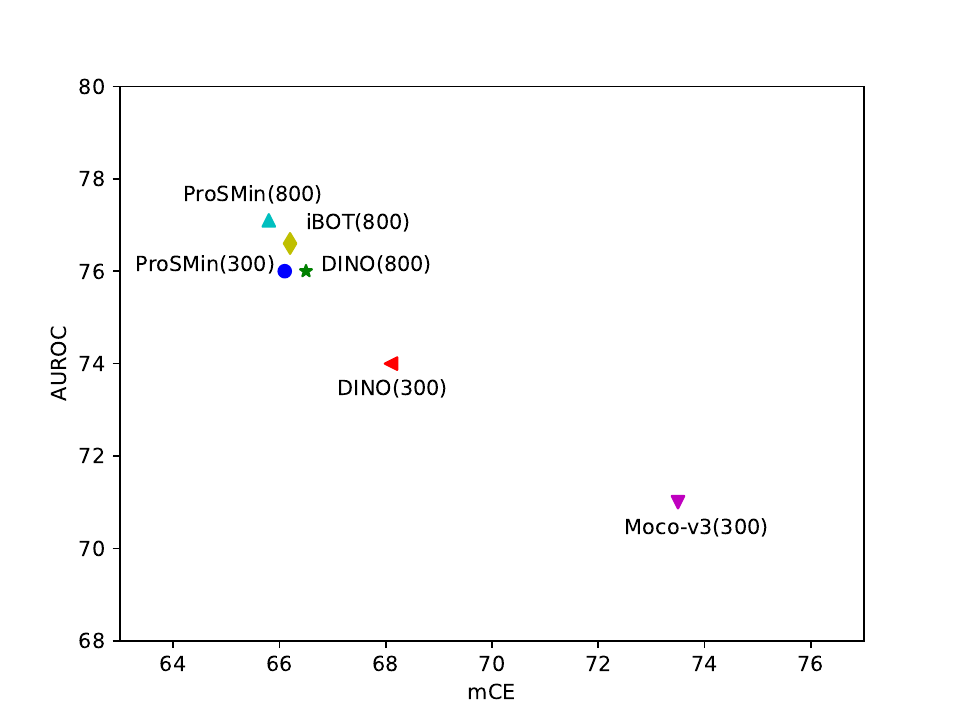}
    \vspace{-10pt}
    \caption{\textbf{OOD detection and corrupted dataset evaluation}, Methods with higher AUROC and lower mCE are better. Our method and iBOT have the best performance across both evaluations. }\label{fig:ood-corrupted}
\end{figure}

\textbf{Out-of-distribution detection ~}
The ability of a model to recognize test samples from classes that were not present during training is evaluated using OOD detection, as discussed in \cite{geng2020recent}. We conduct experiments on ImageNet-O~\cite{srivastava2022out} to assess the generalization of the model from IND to OOD datasets, as well as to predict the uncertainty of the models on OOD datasets. Note that  evaluation is performed directly after self-supervised pretraining without a fine-tuning step. The y-axis of Figure~\ref{fig:ood-corrupted} shows and compares the results for the OOD task in terms of AUROC. Remarkably, our method, trained for 300 epochs, exhibits competitive performance comparable to iBOT, which was trained for 800 epochs. This outcome aligns with our expectations as we directly use the probabilistic latent representation for this task. 

\begin{table}[H]
\caption{\textbf{IND Generalization (or Linear Evaluation)}: Top-1 accuracy, $\kappa$-NN, ECE, and NLL averaged over in-distribution on test samples of the \textbf{ImageNet} dataset where the encoder is \textit{ViT-S/16} over 800 epochs. The best score for each metric is shown in \textbf{bold}, and the second-best is \underline{underlined}. }
\label{table:calibration:imagenet}
  \centering
  \scalebox{.99}{
  \begin{tabular}{l c c c c }
   \toprule
       Method    & \multicolumn{1}{c}{Top-1 Acc (\%)  ($\uparrow$)} & \multicolumn{1}{c}{$\kappa$-NN  ($\uparrow$)}  & 
       \multicolumn{1}{c}{NLL  ($\downarrow$)} &
       \multicolumn{1}{c}{ECE  ($\downarrow$)}  \\ 
    \midrule
    DINO~\cite{caron2021emerging}   & 76.8 & {74.5} & {0.919} & 0.015  \\
    MOCO-V3~\cite{chen2021empirical}& 73.2 & 64.7 & 1.152 & 0.027  \\
    i-BOT~\cite{zhou2021ibot}  &\underline{77.9} & \underline{75.2} & \underline{0.918} & \underline{0.013}  \\
    \rowcolor{cyan!10}
    ProSMin & \textbf{78.4} & \textbf{76.2} & \textbf{0.900} & \textbf{0.006}  \\      
    \bottomrule
  \end{tabular}
  }
\end{table}

\textbf{Corrupted dataset evaluation ~}
An essential aspect of model robustness is its capacity to produce precise predictions when the test data distribution changes. We examine model robustness under the context of \emph{covariate shift}. Figure~\ref{fig:ood-corrupted} presents the improved performance metrics. Our method achieves better results than the baseline and comparable predictive performance to the baseline in terms of mCE.

\textbf{ Semi-supervised and Low-shot learning on ImageNet ~}

Following the semi-supervised protocol established in~\cite{chen2020simple}, we employ fixed 1\% and 10\% splits of labeled training data from ImageNet. In Table~\ref{table:semisupervised}, we compare our performance against several concurrent models, including the baseline (DINO). Our results demonstrate that our features trained for 300 epochs are on par with DINO. Moreover, we achieve better results for 10\% compared to state-of-the-art methods despite our network being pretrained for significantly less time (300 epochs versus 800 epochs).

We assess our model's efficacy on a low-shot image classification task, where we train logistic regression on frozen weights with 1\% and 10\% labels. It's important to note that this experiment was performed on frozen weights, without finetuning. Table~\ref{table:semisupervised} shows our features are on par with state-of-the-art semi-supervised models which these methods trained for 800 epochs. 


\begin{table*}
\caption{\small \textbf{Low-shot and semi-supervised evaluation}: Top-1 accuracy (ACC), ECE, and NLL for semi-supervised on ImageNet classification using 1\% and 10\% training examples fine-tuning. and Low-shot results with 
frozen ViT features.}
\label{table:semisupervised}
  \centering
  \scalebox{.9}{
  \begin{tabular}{l c c c c }
   \toprule
       Method    & \multicolumn{1}{c}{ 1\%} & \multicolumn{1}{c}{ 10\%} & Architecture & Parameters \\

    \midrule   
Semi-supervised &  &  &  &  \\   
DINO~\cite{caron2021emerging} & 60.3 & 74.3 & ViT-S/16 &  21 \\     
i-BOT~\cite{zhou2021ibot} & \underline{61.9} & \underline{75.1} & ViT-S/16 & 21\\ 
\rowcolor{cyan!10}
ProSMin (ours) & \textbf{62.1} & \textbf{75.6} & ViT-S/16  & 21 \\
    \midrule
Low-shot learning &  &  &  &   \\
DINO~\cite{caron2021emerging} & {64.5} & 72.2 & ViT-S/16  & 21 \\      
i-BOT~\cite{zhou2021ibot} & \underline{65.9} & \underline{73.4} & ViT-S/16 & 21 \\
 
\rowcolor{cyan!10}
ProSMin (ours) & \textbf{66.1} & \textbf{73.8}  & ViT-S/16  & 21 \\  
    \bottomrule
  \end{tabular}
  }
\end{table*}

\textbf{Transfer learning evaluation ~}
We further assess the generalization capacity of the learned representation on learning a new dataset. We followed the same transfer learning protocol explained in~\cite{caron2021emerging}. To this end, we evaluate the performance of our model pretrained on ImageNet~\cite{deng2009imagenet} to CIFAR10/100~\cite{Krizhevsky2009learning}, an imbalanced naturalist dataset (iNat-18)~\cite{van2018inaturalist}, and the Flower dataset~\cite{nilsback2008automated}. According to the results shown in Table~\ref{table:transfer}, we observe that our method provides a robust solution when transferring to the new dataset. 


\begin{table*}
\caption{\small \textbf{Transfer to new dataset evaluation}: Transfer learning by finetuning pretrained models on different datasets. We report top-1 accuracy. Self-supervised pretraining with ProSMin transfers better than supervised pretraining.}
\label{table:transfer}
  \centering
  \scalebox{.85}{
  \begin{tabular}{l c c c c }
   \toprule
       Method    & CIFAR-10 & CIFAR-100 & iNat-18 & Flowers \\

    \midrule   
DINO~\cite{caron2021emerging}  & 99.0 & 90.5 & 72 & 98.5 \\     
i-BOT~\cite{zhou2021ibot}  & 99.1 & 90.7 & 73.7 & 98.6 \\   
\rowcolor{cyan!10}
ProSMin (ours)    &99.0&90.2  &72.5 &98.5  \\
    \bottomrule
  \end{tabular}
  }
\end{table*}

\section{Ablation study} \label{ablation}

To gain a deeper understanding of the behavior and performance of our proposed method, we conduct several ablation studies to explore various aspects of our approach. Specifically, we investigate the following factors: different scoring rules as an objective function, the hyperparameter of our loss function ($\lambda$), the number of samples used for generating latent representations, the size of the embedding, the effect of the momentum hyperparameter, the impact BN, and prediction network (PL). These investigations aim to provide insights and intuition regarding our approach.

\begin{figure}[!t]
  \centering
  \subfloat[]{\includegraphics[width=0.45\textwidth]{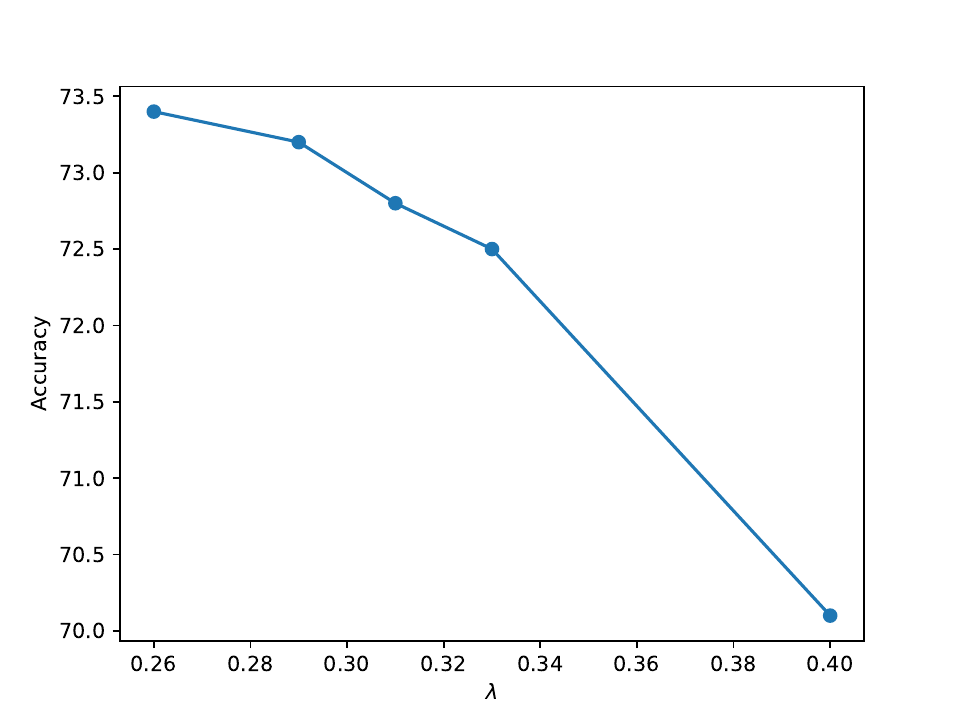}\label{fig:ablation:lambda}}
  \centering
  \subfloat[]{\includegraphics[width=0.45\textwidth]{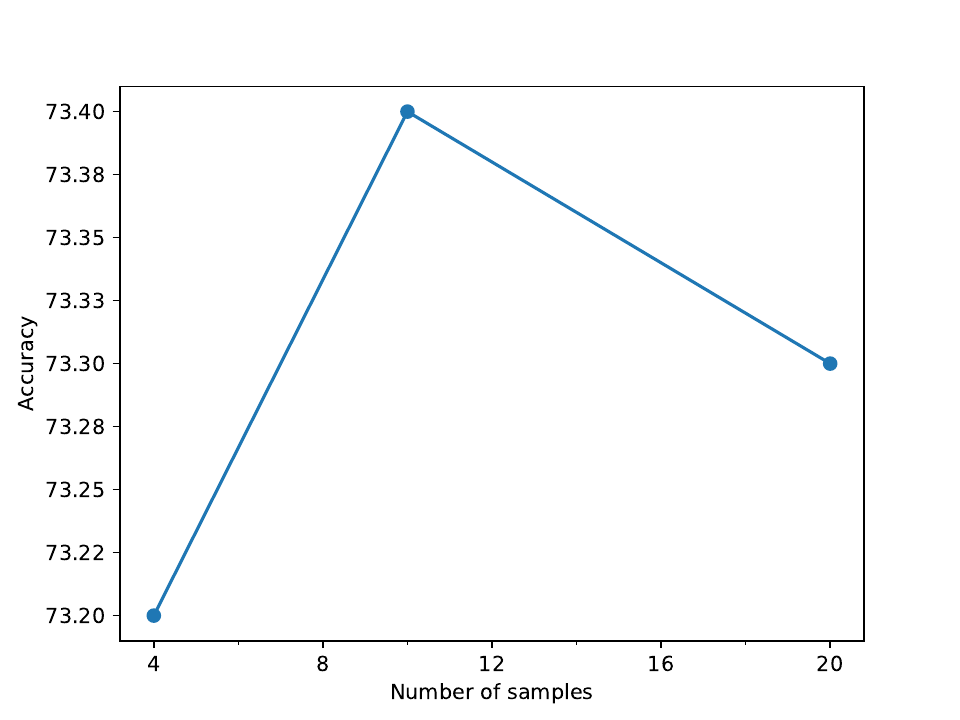}\label{fig:ablation:nsamples}}
  \hfill
  \centering
  \subfloat[]{\includegraphics[width=0.45\textwidth]{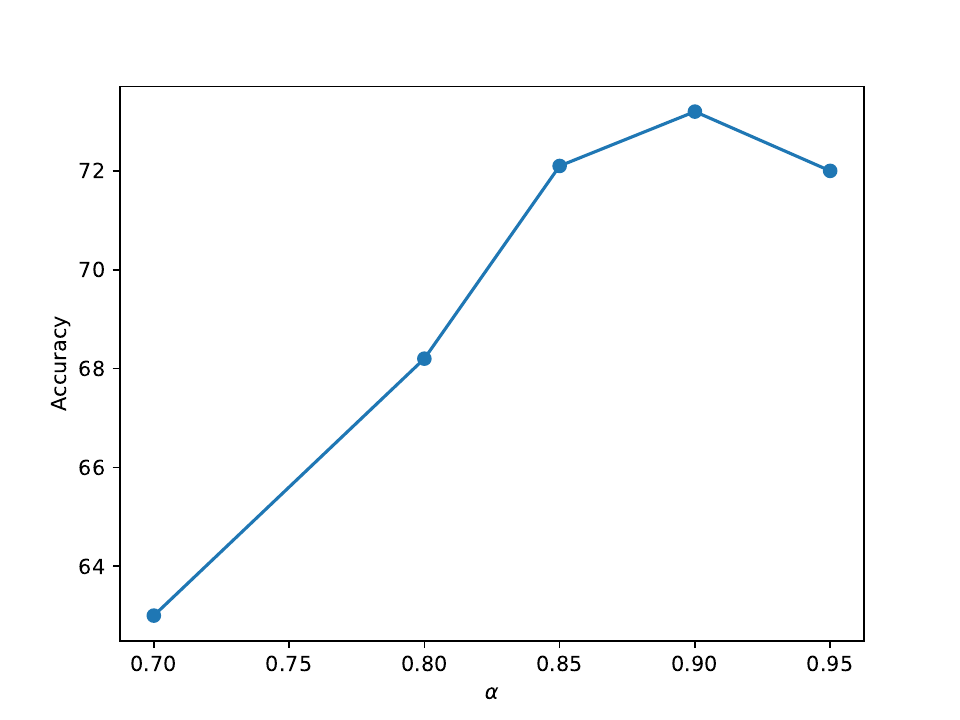}\label{fig:ablation:momentum}}
  \centering
  \subfloat[]{\includegraphics[width=0.45\textwidth]{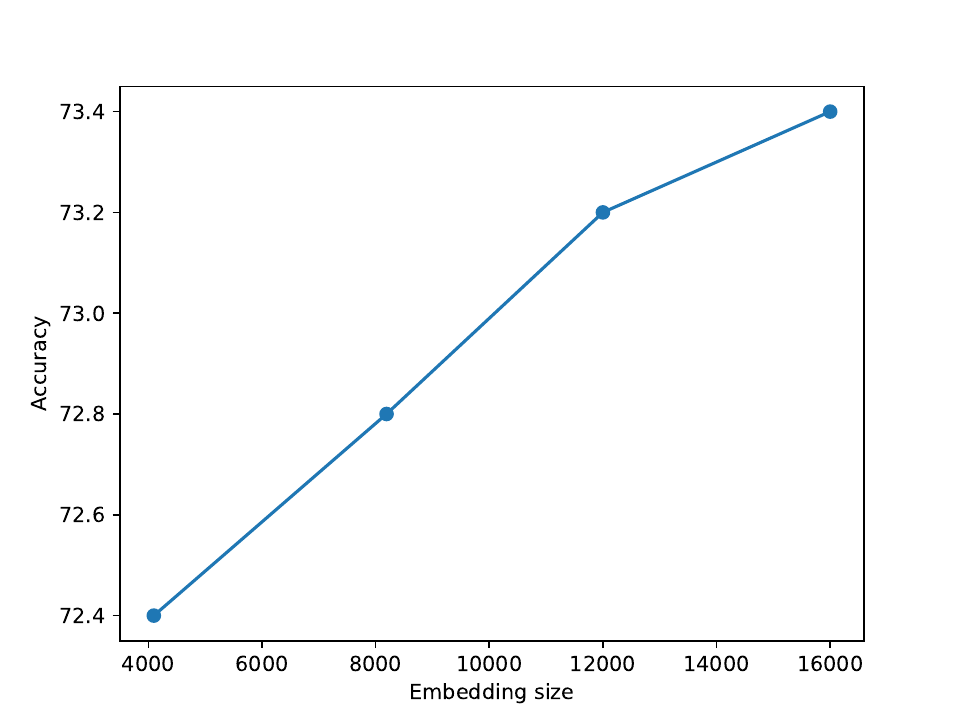}\label{fig:ablation:embedding}}
  \vspace{-2pt}
  \caption{Study of hyperparameters of our proposed ProSMin (a) $\lambda$, (b) Number of samples, (c) Momentum coefficient ($\alpha$), and (d) Size of embedding obtained by 100 epochs on ImageNet dataset.}
\label{fig:ablation:study}
\end{figure}


\textbf{Impact of scoring rule~} We conduct a series of experiments to explore alternative scoring rules, including kernel scoring rules and various variations of energy scoring rules, for our objective function. The results, as presented in Table~\ref{table:ablation}, demonstrate that the kernel scoring rule exhibits instability. Since the kernel score becomes negative and it is not anymore a proper scoring rule (See our proof in Section.~\ref{sec:method:proof}). However, the energy score with $\beta = 1$ ($L1$ loss) yields the best performance among the tested scoring rules. We provide a theoretical explanation for $L1$ and $L2$ (see Section~\ref{sec:appendix:energyloss}).

\begin{table*}
\caption{\small \textbf{Important component for training}. We investigate the effect of each component on the Linear evaluation performance. The first line shows the best combination. PL. is the prediction layer. BN. is Batch normalization layer}
\label{table:ablation}
  \centering
  \scalebox{0.9}{
  \begin{tabular}{l c c c c c }
   \toprule
    Energy (L1 loss) & Kernel score & Energy (L2-loss) & BN. & PL.  & Accuracy \\
    \midrule 
    \rowcolor{cyan!10}
  \cmark & \xmark & \xmark& \xmark  & \cmark &  73.2 \\   
  \xmark & \cmark & \xmark& \xmark  & \cmark &  1.1 \\   
  \xmark & \xmark & \cmark& \xmark  & \cmark &  43 \\ 
  \cmark & \xmark & \xmark& \cmark  & \cmark &  68.5 \\
  \cmark & \xmark & \xmark& \xmark  & \xmark &  71.2 \\
  \cmark & \xmark & \xmark& \xmark  & \cmark &  71 \\
    \bottomrule
  \end{tabular}
  }
\end{table*}


\begin{table}
\caption{\small \textbf{Evaluation of Computational Efficiency} We conduct a thorough analysis of the computational efficiency of our novel probabilistic approach in comparison to alternative self-supervised methods. This evaluation encompasses memory utilization and computational expenditure.}
\label{table:computation}
  \centering
  \scalebox{.9}{
  \begin{tabular}{l c c c c c }
   \toprule
       Method (ViT-s)   & parameters (M) & im/s & time/ 300-epochs (hr) & number of GPUs & memory (G)\\

    \midrule   
DINO~\cite{caron2021emerging}  & 21 & 1007 & 72.6 & 16 & 15.4 \\ 
i-BOT~\cite{zhou2021ibot}  & 21 & 1007 & 73.8 & 16 & 19.5 \\   
\rowcolor{cyan!10}
ProSMin (ours) & 21 & 1007  & 98 & 8 & 21.1  \\
    \bottomrule
  \end{tabular}
  }
\end{table}

\textbf{Study of hyperparameters~}
Figure~\ref{fig:ablation:lambda} depicts the influence of the hyperparameter $\lambda$ utilized in our proposed loss function, which controls the impact of $S^{1}(P_{\bm{\theta}}, \bm{z}_{\bm{\xi}})$ and $S^{2}(P_{\bm{\theta}}, \bm{z}_{\bm{\xi}})$ on the objective function. 

An ablation analysis was conducted to investigate the effect of increasing the number of samples from $q_{\bm{\theta}}$, as illustrated in Figure~\ref{fig:ablation:nsamples}. The results demonstrate that employing four samples yields satisfactory performance.

Figure~\ref{fig:ablation:momentum} showcases the outcomes obtained from the knowledge distillation rate. In previous approaches, the exponential moving average parameter initiated from a value relatively close to 1 (e.g., 0.996 \cite{grill2020bootstrap}, \cite{caron2021emerging}). However, in our case, $\alpha$ commences from 0.9, implying a faster pace of knowledge distillation.

Furthermore, we examined the impact of different sizes for the embedding vector, as presented in Fig.~\ref{fig:ablation:embedding}. The results obtained after 100 epochs reveal that increasing the embedding size leads to improved performance. However, it should be noted that larger embedding sizes necessitate additional computational resources, thus our choice of size was based on the available computational capacity.

Table~\ref{table:ablation} provides insights into the influence of batch normalization in the prevention of representation collapse \cite{grill2020bootstrap}. As our framework operates in a probabilistic setting, the inclusion of batch normalization is unnecessary for averting collapse. Additionally, the prediction layer (PL $q_{\theta}$) enhances performance by facilitating improved feature extraction in online networks.

\textbf{Analysis of computational cost}
We further assess the effectiveness of our proposed approach and compare it with DINO and iBOT in Table~\ref{table:computation}. The presented values are obtained from the data reported in the DINO and iBOT papers. However, these papers do not include information regarding the total number of parameters.

\section{Conclusion} \label{conclusion}
\vspace{-5pt}
In this paper, we presented \emph{ProSMin} as a novel probabilistic self-supervised framework that involves two neural networks which collaborate and learn from each other using an augmented format. Our framework is trained by minimizing a proposed scoring rule. We provided theoretical justification and showed that our modified loss is strictly proper. We evaluated ProSMin across different tasks, including in-distribution generalization, out-of-distribution detection, dataset corruption, transfer learning, and semi-supervised learning. The results demonstrate that our method achieves superior performance in terms of accuracy and calibration, thus showing the effectiveness of our proposed approach.

\paragraph{Broader impact and limitations}

This study has the potential to inspire new algorithms and stimulate theoretical and experimental exploration. 
The algorithm presented here can be used for many different probabilistic downstream tasks, including (but not limited to) uncertainty quantification, density estimation, image retrieval, probabilistic unsupervised clustering, program debugging, image generation, music analysis, and ranking.
In addition, we believe that our extended concept probabilistic framework opens many interesting avenues for future development in self-supervised learning, and addresses many problems of existing models, such as avoiding representation collapse.

However, there are several limitations. One limitation of our model compared to other learning methods (such as supervised learning) is that self-supervised learning may require more computational resources and training time. However, considering that our proposed method does not require manual annotation, which is usually very expensive, we would argue that this trade-off is acceptable. In addition, due to limited computational resources, we provide results for fewer epochs compared to other methods. Furthermore, we believe that the results can be improved with extensive hyperparameter optimization.



{
\small
\bibliography{neurips_2023}
}

\newpage
\section{Scoring Rules}~\label{sec:scoring_rule}
The notation employed in this study aligns with the work of Gneiting  et al. \cite{gneiting2007strictly} regarding scoring rules $S(P, \bm{x})$, where $P$ represents the predictive distribution and $\bm{x}$ denotes the observed data. By assuming that $\mathbf{X}$ follows a true distribution $Q$.
With this assumption, we first explain different variations of scoring rules, such as the expected scoring rule~\ref{sec:app:sc:ex}, the proper scoring rule~\ref{sec:app:sc:proper}, the energy scoring rule~\ref{sec:energy}, and the kernel scoring rule~\ref{sec:kernel}. 

We then discuss the unbiased estimation properties of the scoring rule~\ref{sec:app:unbiased} for kernel score~\ref{sec:app:kernel} and energy score~\ref{sec:app:energy}, which we used as the objective function in our study.

In subsection \ref{bem_exchange}, we establish the interchangeability of the expectation and gradient in our theoretical derivative, enabling us to derive gradients in a conventional manner. This is crucial due to the incorporation of samples within the scoring rule and the utilization of non-differential activation functions. Subsequently, in subsection \ref{unb_est}, we introduce an unbiased estimate of the gradient. By completing these three steps, we guarantee effective convergence of our algorithm (refer to \ref{sec:method:proof}).

\subsection{Expected scoring rule}~\label{sec:app:sc:ex}
We define the scoring rule $S(P,x)$ as a function of the distribution $P$ and the observation $x$ of $\mathbf{X}$.
The expected scoring rule is defined as:
    \begin{align}
        S(P, Q):=\mathbb{E}_{\mathbf{X} \sim Q} S(P, \mathbf{X}).
    \end{align}
    
\subsection{Proper scoring rule}~\label{sec:app:sc:proper}
A scoring rule $S$ is called \emph{proper} w.r.t a set of distributions $\mathcal{P}$, if for all $P, Q \in \mathcal{P}$ the expected score $S(P, Q)$ is minimized in $Q$ at $Q = P$. 
A scoring rule $S$ is called \textit{strictly proper} if there exists the unique minimum $S(P, Q) > S(Q, Q)$ for $Q \neq P$.

\subsection{Energy score}\label{sec:energy}
The energy score is defined by:
\begin{align}
S_{\mathrm{E}}^{(\beta)}(P, \mathbf{x})=2 \cdot \mathbb{E}\left[\|\tilde{\mathbf{X}}-\mathbf{x}\|_2^\beta\right]-\mathbb{E}\left[\left\|\tilde{\mathbf{X}}-\tilde{\mathbf{X}}^{\prime}\right\|_2^\beta\right], \quad \tilde{\mathbf{X}} \independent \tilde{\mathbf{X}}^{\prime} \sim P,
\end{align}
 where $\beta \in (0,2)$. 
 It is a strictly proper scoring rule for the class of probability measures $\mathcal{P}$ such that $\mathbb{E}_{\tilde{\mathbf{X}} \sim P}\|\tilde{\mathbf{X}}\|^\beta<\infty$. 
 An unbiased estimate can be obtained by replacing the expectations in $S_{\mathrm{E}}^{(\beta)}$ with empirical means over draws from $P$.

\subsection{Kernel score}\label{sec:kernel}
Let $k(\cdot, \cdot)$ be a positive definite kernel. 
The kernel score for $k$ is defined by
    \begin{align}
    S_k(P, \mathbf{x})=\mathbb{E}\left[k\left(\tilde{\mathbf{X}}, \tilde{\mathbf{X}}^{\prime}\right)\right]-2 \cdot \mathbb{E}[k(\tilde{\mathbf{X}}, \mathbf{x})], \quad \tilde{\mathbf{X}} \independent \tilde{\mathbf{X}}^{\prime} \sim P.
    \end{align}
The kernel score is a proper scoring rule for the class of probability distribution $P$ for which holds, that $\mathbb{E}_{\tilde{\mathbf{X}}, \tilde{\mathbf{X}}^{\prime} \sim P}\left[k\left(\tilde{\mathbf{X}}, \tilde{\mathbf{X}}^{\prime}\right)\right] < \infty$.
Under the condition that the kernel Maximum Mean Discrepancy is a metric, the kernel score is strictly proper. 
This condition is satisfied by the Gaussian kernel, which will be used in this work. 
Let $\gamma$ be a scalar bandwidth. We define the Gaussian kernel as follows:
\begin{align*}
    k(\tilde{\mathbf{x}}, \mathbf{x})=\exp \left(-\frac{\|\tilde{\mathbf{x}}-\mathbf{x}\|_2^2}{2 \gamma^2}\right).
\end{align*}

We get an unbiased estimate by replacing the expectations in $S_{\mathrm{k}}$ with the empirical means overdraws from $P$.

\subsection{Unbiased estimators of the expected scoring rule}~\label{sec:app:unbiased}
Let $N$ be the batch size and $r$ be the number of samples drawn from the online-induced distribution.
Let $z_\theta = (z_{1},...,z_{r})$ be the sample drawn from the online.
We recall the definition of the Expected Scoring Rule:
\begin{align*}
    S(P_\theta, P_\xi):=\mathbb{E}_{z_\xi \sim P_\xi} [S(P_\theta, z_\xi)],
\end{align*}
where $z_\xi$ denotes the output of the target network while $P_\theta$ represents the online induced multivariate normal distribution.

\subsubsection{Kernel score}~\label{sec:app:kernel}
We can write the estimated kernel score as follows:
\begin{align*}
    \hat{S}(P_\theta, P_\xi) = 
    \dfrac{1}{N} \sum_{i=1}^{N} \!\left[ \frac{1}{r(r-1)} \sum_{j,k = 1, j \neq k} k(z_{j}^{i}, z_{k}^{i}) -
    \dfrac{2}{r} \sum_{j=1}^{r} k(z_{j}^{i}, z_\xi^{i})\!\right]
\end{align*}

\subsubsection{Energy score}~\label{sec:app:energy}
Consider $\beta \in (0,2)$ and $N$ be the batch size.
The Estimated Energy Score can be written as:
\begin{align*}
    \hat{S}(P_\theta, P_\xi) = 
    \dfrac{1}{N} \sum_{i=1}^{N} \!\left[ 
    \dfrac{2}{r} \sum_{j=1}^{r} \Vert z_{j}^{i} - z_\xi^{i}\Vert^{\beta}_{2} -
    \frac{1}{r(r-1)} \sum_{j,k = 1, j \neq k} \Vert z_{j}^{i} - z_{k}^{i} \Vert^{\beta}_{2} \!\right]
\end{align*}

\subsubsection{Partial derivatives of the estimated energy score}~\label{sec:appendix:energyloss}
Let $\beta \in (0,2)$ and let $N$ be the batch size.
    Let $x_{i}$ be the input data for $i \in {1,...,N}$.
    Let $r \in \mathbb{N}$ be the number of samples.
    Let $\mu_{i}=f^{(1)}_{\theta}(x_{i})$ and let $\sigma^{2}_{i}=f^{(2)}_{\theta}(x_{i})$, where $f^{(j)}$ denotes the separated branches for mean and variance in the last layer of the online network for $j \in {1,2}$.
    We can calculate the partial derivatives needed for stochastic gradient descent as follows. Then, the formula for the mean is: 
    \begin{align*}
        \dfrac{\partial}{\partial \mu}\hat{S}(P_\theta, P_\xi) &= 
        \dfrac{\partial}{\partial \mu} \dfrac{1}{N} \sum_{i=1}^{N} \left[ 
        \dfrac{2}{r} \sum_{j=1}^{r} \left\Vert \mu_{i} + \epsilon_{j} \sqrt{\sigma^{2}_{i}} - \phi_\xi^{i} \right\Vert^{\beta}_{2} \right. \\
        & \quad \left. - \frac{1}{r(r-1)} \sum_{j,k = 1, j \neq k} \left\Vert \mu_{i} + \epsilon_{j} \sqrt{\sigma^{2}_{i}} - \mu_{i} + \epsilon_{k} \sqrt{\sigma^{2}_{i}} \right\Vert^{\beta}_{2} \right] \\
        &= \dfrac{1}{N} \sum_{i=1}^{N} \dfrac{2}{r} \sum_{j=1}^{r} \beta \left\Vert \mu_{i} + \epsilon_{j} \sqrt{\sigma^{2}_{i}} - \phi_\xi^{i} \right\Vert^{\beta-1}_{2} \dfrac{\partial}{\partial \mu} \left[\mu_{i} + \epsilon_{j} \sqrt{\sigma^{2}_{i}} -\phi_\xi^{i} \right] \\
        &= \dfrac{1}{N} \sum_{i=1}^{N} \dfrac{2}{r} \sum_{j=1}^{r} \beta \left\Vert f^{(1)}_{\theta}(x_{i}) + \epsilon_{j} \sqrt{f^{(2)}_{\theta}(x_{i})} - \phi_\xi^{i} \right\Vert^{\beta-1}_{2} \nabla_{\theta} f^{(1)}_{\theta}(x_{i}) \\
    \end{align*}
    
    while the formula for the variance is:
      \begin{align*}
        \dfrac{\partial}{\partial \sigma^{2}}\hat{S}(P_\theta, P_\xi) &= 
        \dfrac{\partial}{\partial \sigma^{2}} \dfrac{1}{N} \sum_{i=1}^{N} \left[ 
        \dfrac{2}{r} \sum_{j=1}^{r} \left\Vert \mu_{i} + \epsilon_{j} \sqrt{\sigma^{2}_{i}} - \phi_\xi^{i} \right\Vert^{\beta}_{2} \right. \\
        & \quad \left. - \frac{1}{r(r-1)} \sum_{j,k = 1, j \neq k} \left\Vert \mu_{i} + \epsilon_{j} \sqrt{\sigma^{2}_{i}} - \mu_{i} + \epsilon_{k} \sqrt{\sigma^{2}_{i}} \right\Vert^{\beta}_{2} \right] \\
        &= \left[  \dfrac{\beta}{N} \sum_{i=1}^{N} \left[ \dfrac{2}{r} \sum_{j=1}^{r} \left\Vert f^{(1)}_{\theta}(x_{i}) + \epsilon_{j} \sqrt{f^{(2)}_{\theta}(x_{i})} - \phi_\xi^{i} \right\Vert^{\beta-1}_{2}  \right. \right. \\
        & \quad \left. \left. - \frac{1}{r(r-1)} \sum_{j,k = 1, j \neq k} \left\Vert  \sqrt{f^{(2)}_{\theta}(x_{i})}\left(\epsilon_{j} - \epsilon_{k} \right)  \right\Vert^{\beta-1}_{2}\right] \right]\nabla_{\theta} f^{(2)}_{\theta}(x_{i}) \\
    \end{align*}

\subsubsection{Exchangeability of gradient and expectation}\label{bem_exchange}

This proof follows a similar argumentation as the proof performed in \cite{pacchiardi2022likelihood}.
We want to solve
\begin{equation}
    \hat{\bm{\theta}}:=\argmin_{\bm{\theta}} J(\bm{\theta}), \quad J(\bm{\theta}) = \dfrac{1}{N} \sum_{i=1}^{N} S(P_{\bm{\theta}}, \bm{z}^i_{\bm{\xi}})
\end{equation}
Finding a minimum in our network architecture is done via stochastic gradient descent (SGD) or an algorithm exploiting the same properties as SGD.
Recall that the scoring rule is defined as an expectation over samples from $P_{\bm{\theta}}$.
Additionally, we can describe the scoring rule as some function $h$ of independent inputs $\bm{z}^i_j \independent \bm{z}^i_k$.
Hence we can write:
\begin{equation}
        S(P_{\bm{\theta}}, \bm{z}^i_{\bm{\xi}})=\mathbb{E}_{\bm{z}^i_j,\bm{z}^i_k \sim P_{\bm{\theta}}} \left[h \left(\bm{z}^i_j,\bm{z}^i_k,\bm{z}^i_{\bm{\xi}}\right) \right]
\end{equation}

Let $\mu_{i}, \sigma^{2}_{i}$ be the estimated mean and variance, respectively, of the online network $f_{\bm{\theta}}$ for the $i$-th input sample $\bm{x}_{i}$.
With $P_{\bm{\theta}}$ being the distribution induced by the online network and $g_{\bm{\theta}}$ being its transformation, we use the reparametrization trick and get
\begin{equation}
    J(\bm{\theta})= \dfrac{1}{N} \sum_{i=1}^{N} \mathbb{E}_{\epsilon^i_j,\epsilon^i_k\sim \mathcal N(0,1)} \left[h ( g_{\bm{\theta}}(\epsilon^i_j,\mu_{i}, \sigma^{2}_{i}), g_{\bm{\theta}}(\epsilon^i_k,\mu_{i}, \sigma^{2}_{i}), \bm{z}^i_{\bm{\xi}} ) \right]
\end{equation}
We now can write the derivative as follows:
\begin{align*}
        \nabla_{\bm{\theta}} J(\bm{\theta}) &= \nabla_{\bm{\theta}} \dfrac{1}{N} \sum_{i=1}^{N} \mathbb{E}_{\epsilon^i_j,\epsilon^i_k\sim \mathcal N(0,1)} \left[h ( g_{\bm{\theta}}(\epsilon^i_j,\mu_{i}, \sigma^{2}_{i}), g_{\bm{\theta}}(\epsilon^i_k,\mu_{i}, \sigma^{2}_{i}), \bm{z}^i_{\bm{\xi}} ) \right] \\
    &= \dfrac{1}{n} \sum_{i=1}^{n} \mathbb{E}_{\epsilon^i_j,\epsilon^i_k\sim \mathcal N(0,1)} \left[\nabla_{\bm{\theta}} h ( g_{\bm{\theta}}(\epsilon^i_j,\mu_{i}, \sigma^{2}_{i}), g_{\bm{\theta}}(\epsilon^i_k,\mu_{i}, \sigma^{2}_{i}), \bm{z}^i_{\bm{\xi}} ) \right]
\end{align*}
The exchange between expectation and gradient is not trivial because of the non-differentiability of functions (e.g., ReLU) within the network function $g_{\bm{\theta}}$.
We can still perform this step by using Theorem 5 from \cite{pacchiardi2022likelihood}, considering mild conditions on the neural network architecture of $g_{\bm{\theta}}$.

\subsubsection{Unbiased estimate of the gradient}\label{unb_est}
Let $\mathcal{B}$ be a random subset (batch) of the data set. Using \ref{bem_exchange}, we can obtain unbiased estimates of $\nabla_{\bm{\theta}} J(\bm{\theta})$ using samples $\epsilon^j_i\sim \mathcal N(0,1), j = 1,...,r$ for each $i \in \{1,...,N\}$:
\begin{equation}
    \widehat{\nabla_{\bm{\theta}} J(\bm{\theta})} = \dfrac{1}{|\mathcal{B}|} \sum_{i \in \mathcal{B}} \dfrac{1}{r(r-1)} \sum_{j,k=1, j \neq k}^{r} \nabla_{\bm{\theta}} h ( g_{\bm{\theta}}(\epsilon^k_i,\mu_{i}, \sigma^{2}_{i}), g_{\bm{\theta}}(\epsilon^j_i,\mu_{i}, \sigma^{2}_{i}), \bm{z}^i_{\bm{\xi}} )
\end{equation}

\subsubsection{Detailed implementation}\label{Implementation}


In addition to the detail in Section~\ref{sec:implementation}, it's important to mention that the target network takes two global augmented samples while the online network takes two global augmented samples and 16 local augmentations for multi-cropping samples. Then, the  $r$ in Eq.~\ref{eq:2} refers to the number of augmentation samples multiplied by the number of samples. Furthermore, in Eq.~\ref{eq:2} $z_{\xi}$ denotes one of the global augmented samples.


\end{document}